\documentclass[10pt,letterpaper, table]{article}
\PassOptionsToPackage{table}{xcolor}

\usepackage{cogsci}

\cogscifinalcopy %

\usepackage{pslatex}
\usepackage{apacite}
\usepackage{float} %

\usepackage{xcolor}
\usepackage{graphicx}
\usepackage{ascmac} 
\usepackage{url}
\usepackage{enumitem}

\usepackage{array} %
\usepackage{multirow} %
\usepackage{natbib}
\usepackage{booktabs}
\usepackage[export]{adjustbox}   %

\usepackage[multiple]{footmisc}

\makeatletter
\def\new@fontshape{}
\makeatother
\usepackage{gb4e}
\noautomath

\definecolor{myblue}{HTML}{3B78D9}

\usepackage{leipzig}

\title{Bridging Perception and Language: A Systematic Benchmark for LVLMs’ Understanding of Amodal Completion Reports}

\author{%
    Amane Watahiki${}^{1}$ Tomoki Doi${}^{1}$ Taiga Shinozaki${}^{2,1}$ \\
    {\bf Satoshi Nishida}${}^{3,4,5,6}$ {\bf Takuya Niikawa}${}^{7}$ {\bf Katsunori Miyahara}${}^{5}$ {\bf Hitomi Yanaka}${}^{1}$\\
    ${}^{1}$The University of Tokyo \quad ${}^{2}$Keio University \quad ${}^{3}$NICT \\
    ${}^{4}$Osaka University \quad ${}^{5}$Hokkaido University \quad ${}^{6}$CiNET \quad ${}^{7}$Kobe University\\
    \texttt{amanew@g.ecc.u-tokyo.ac.jp, snzktig@keio.jp, s-nishida@nict.go.jp}\\
    \texttt{\{doi-tomoki701, hyanaka\}@g.ecc.u-tokyo.ac.jp}, \texttt{niitaku11@gmail.com, kmiyahara@chain.hokudai.ac.jp}
}

\begin{document}

\maketitle

\begin{abstract}
One of the main objectives in developing large vision-language models (LVLMs) is to engineer systems that can assist humans with multimodal tasks, including interpreting descriptions of perceptual experiences. A central phenomenon in this context is amodal completion, in which people perceive objects even when parts of those objects are hidden. Although numerous studies have assessed whether computer-vision algorithms can detect or reconstruct occluded regions, the inferential abilities of LVLMs on texts related to amodal completion remain unexplored. To address this gap, we constructed a benchmark grounded in Basic Formal Ontology to achieve a systematic classification of amodal completion. Our results indicate that while many LVLMs achieve human-comparable performance overall, their accuracy diverges for certain types of objects being completed. Notably, in certain categories, some LLaVA-NeXT variants and Claude 3.5 Sonnet exhibit lower accuracy on original images compared to blank stimuli lacking visual content. Intriguingly, this disparity emerges only under Japanese prompting, suggesting a deficiency in Japanese-specific linguistic competence among these models.

\end{abstract}

\section{Introduction}

A key goal in developing large vision--language models (LVLMs) is to enable them to support humans in multimodal tasks, which requires accurate interpretation of texts describing perceptual experiences of images. A central phenomenon in human perception is \textit{amodal completion} (AC), in which individuals perceive whole objects even when parts are occluded. For example, a subject might report ``seeing two rectangles'' in Figure~\ref{fig:amodal_example}, despite one being partially hidden. This perceptual ``filling in'' occurs without direct sensory input and so is called ``amodal'' completion.

\begin{small}
\begin{figure}[!t]
 \centering
 \includegraphics[width=0.5\linewidth]{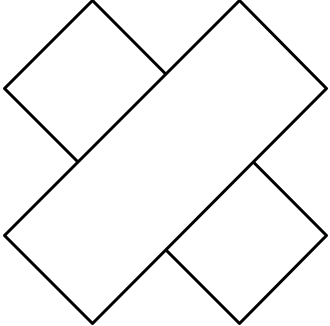}
\caption{Example of amodal completion \citep{Michotte1991-kp}. Although this figure is drawn on a single plane, it appears as if one rectangle is partially hidden under another rectangle, and the perceiver completes the occluded part ``amodally.''}
\label{fig:amodal_example}
\end{figure}
\end{small}

AC is essential in daily life \citep{Nanay2018-vq} and appears frequently in human narratives. Therefore, for LVLMs to handle these narratives, they must interpret AC-related descriptions accurately. However, AC poses unique challenges: although the occluded parts are not physically visible, they appear to be there, leading subjects to use perceptual verbs such as ``looks like.'' These verbs often coexist with the object's invisibility, creating seemingly paradoxical expressions. Humans resolve this paradox naturally, but it remains unclear whether LVLMs can do the same.

Previous studies have examined how models reconstruct occluded object properties such as shape or color, but the focus was mostly on visual processing \citep{AO-2023}, leaving textual inference in AC underexplored. Furthermore, AC was typically classified into only two or three coarse categories, therefore lacking clarity and nuance. Researchers often assume that the type of completed object is obvious, and few provide criteria for consistent tagging. This reliance on intuition hinders standardization and interoperability.

To address the aforementioned issues, we introduce \textbf{VACT} (\textbf{V}isual \textbf{A}modal \textbf{C}ompletion with \textbf{T}exts), a benchmark using a two-choice question--answer format, with categories tagged based on \textbf{Basic Formal Ontology (BFO)}, which provides a structured, interoperable framework for categorizing entities. By comparing model and human performances across these fine-grained categories, we identify where LVLMs struggle. This also serves as a case study demonstrating the utility of ontologies such as BFO in the evaluation of artificial intelligence (AI).

Interestingly, some LLaVA-NeXT models and Claude 3.5 Sonnet perform less well on original images and better on blank ones for some categories, but only in the Japanese setting. We hypothesize that this is due to difficulty in distinguishing the typical versus evidential use of perceptual verbs in Japanese. Evidential use of a perceptual verb indicates the source of the speaker's information, often weakening its original lexical meaning. This allows native Japanese speakers to interpret seemingly paradoxical AC descriptions. While English also marks evidentiality, it does so through different syntactic strategies. This cross-linguistic variation may limit the ability of LVLMs to transfer linguistic knowledge from English to Japanese. Although still a hypothesis, our findings point to a promising direction for future work: bridging perception and language by exploring how evidentiality shapes the expression and interpretation of perceptual experiences across languages.

\section{Related Work}
\label{sec:related}

In previous research, various datasets were constructed to assess the AC capabilities of models in tasks involving \emph{image processing} \citep{AO-2023}. However, none of these tasks assess the inference capabilities of LVLMs with respect to \emph{textual descriptions of AC}, which is the focus of the present study.

Different forms of AC may reveal specific strengths and limitations in LVLM performance. However, the classifications used commonly in previous research are (i) insufficiently fine-grained to yield insights into the specific characteristics of each model and (ii) lack clear definitions for each category. For instance, \citet{AO-2023} identified shape completion, appearance completion, and order perception as key categories in recent AC tasks in image processing. Similarly, other studies classified AC based on whether it involves shape or color completion \citep{Gerbino2020-yk, Pessoa2001-ys}. \citet{van-Lier2015-gq} divided AC into two categories, two-dimensional (2D) and three-dimensional (3D), while \citet{Tse1999-xi} argued that all types of completion can be understood as volume completion. However, as will be shown, the range of objects subject to AC extends beyond these categories. For example, when we see a die, its back is occluded by its front surface, but we still perceive it as a cube. In this case, one could argue that what is being completed is not merely the color or shape but the die as a whole. 

Turning to less-artificial real-world images, it becomes evident that existing categories are too vague and coarse-grained to support a systematic and comprehensive classification scheme. Is the back surface of a drinking glass 2D or 3D? On one hand, one might think that it is a 2D object because what is at stake here is something's ``surface.'' On the other hand, one can also believe that no 2D object can exist because nothing can exist without thickness in the real world. There is nothing wrong with both ways of thinking. In the absence of clear criteria, annotating everyday instances of AC becomes challenging.

\section{Proposed Dataset}
\label{sec:method} %

\subsection{Overview}

In this study, we construct the VACT benchmark dataset, which presents a two-alternative forced choice (2AFC) task requiring models and participants to identify the correct description of an amodally complemted object in an image. Sample images and associated questions from the dataset are shown in Table~\ref{tab:question_updated}. This benchmark is used to evaluate the inference capabilities of LVLMs on text-based AC tasks and to compare their performance against human judgments.

To assess this ability with a finer-grained and more formally defined taxonomy of AC, VACT incorporates a classification framework based on BFO \citep{Arp2015-rt}, which is a top-level ontology designed to structure scientific knowledge systematically across disciplines in a consistent and interoperable fashion. As a top-level ontology, BFO offers high-level categories applicable across all domains of scientific inquiry.

\begin{small} 
\begin{figure}[!t] 
\centering 
\includegraphics[width=1\linewidth]{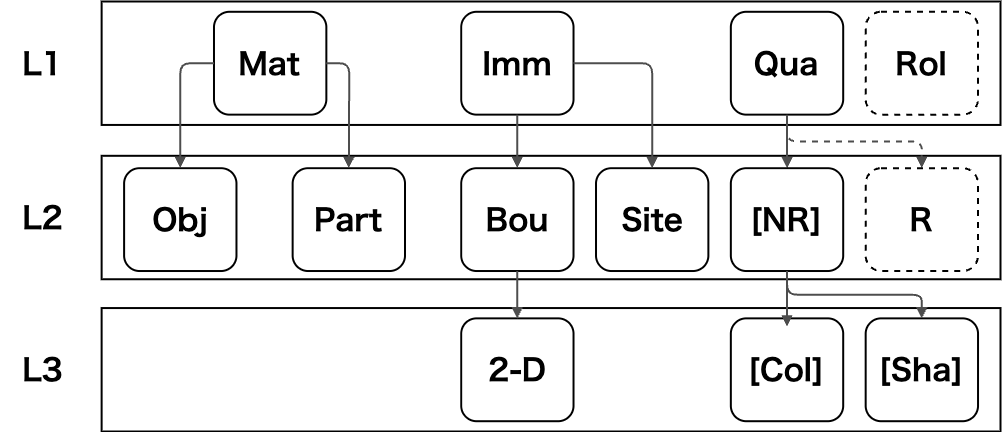} 
\caption{Partial taxonomy of BFO continuant categories. Dotted lines denote categories outside the scope of this study, and those not defined in BFO are enclosed in square brackets ([~]).}
\label{fig:BFO_tree} 
\end{figure} 
\end{small}

Two key advantages motivate the adoption of BFO in this study: (i) BFO is designed to classify all entities into one of its categories, enabling it to accommodate objects and phenomena that do not fit into existing classifications, including those identified by \citet{AO-2023}; (ii) BFO provides a clear definition or elucidation of each BFO category, facilitating systematic classification and reducing ambiguity. 

Ontologies such as BFO classify recurring features of the real world and represent their interrelations through hierarchical taxonomies. Such taxonomies can be formalized as graphs, with nodes denoting categories and edges indicating the subtype or subclass ($is\_a$) relation between these categories. With BFO, we obtain a structured classification of AC as shown in Figure~\ref{fig:BFO_tree}.

\subsection{Details of BFO Categories}

Here, we briefly outline the categories relevant to VACT, with Figure~\ref{fig:BFO_tree} showing the part of the BFO taxonomy that is relevant to our study; see \cite{Arp2015-rt} for a comprehensive treatment of BFO. VACT uses subtypes of both independent and dependent continuants as defined in BFO: independent continuants are entities that exist on their own, while dependent continuants ``inhere'' in or depend upon independent continuants for their existence. This study refers to categories directly under this level as \textbf{first-layer (L1)} categories in our framework. Linked via $is\_a$ relationships, their subcategories are designated as \textbf{second-layer (L2)} or \textbf{third-layer (L3)} categories, depending on their depth in the hierarchy. Figure~\ref{fig:BFO_tree} shows the parent--child relationships among these categories. 

\begin{table*}[htbp]
\caption{Example questions. Although this table shows three texts for one image, VACT consists of two-choice QA tasks; for each image and gold answer, two question pairs are prepared: $(\text{gold}, \text{miscompletion})$ and $(\text{gold, no-completion})$ pairs.}
\label{tab:question_updated}
\centering
\begin{tabular}{|>{\centering\arraybackslash}m{1.9cm}|>{\centering\arraybackslash}m{3.4cm}|>{\centering\arraybackslash}m{3.4cm}|>{\centering\arraybackslash}m{3.4cm}|>{\centering\arraybackslash}m{3.4cm}|}
\hline
\textbf{Question} & \textbf{A} & \textbf{B} & \textbf{C} & \textbf{D} \\ \hline
\textbf{\small{Image}} & 
\makebox[\linewidth][c]{\raisebox{-0.1\height}{\includegraphics[width=0.9\linewidth]{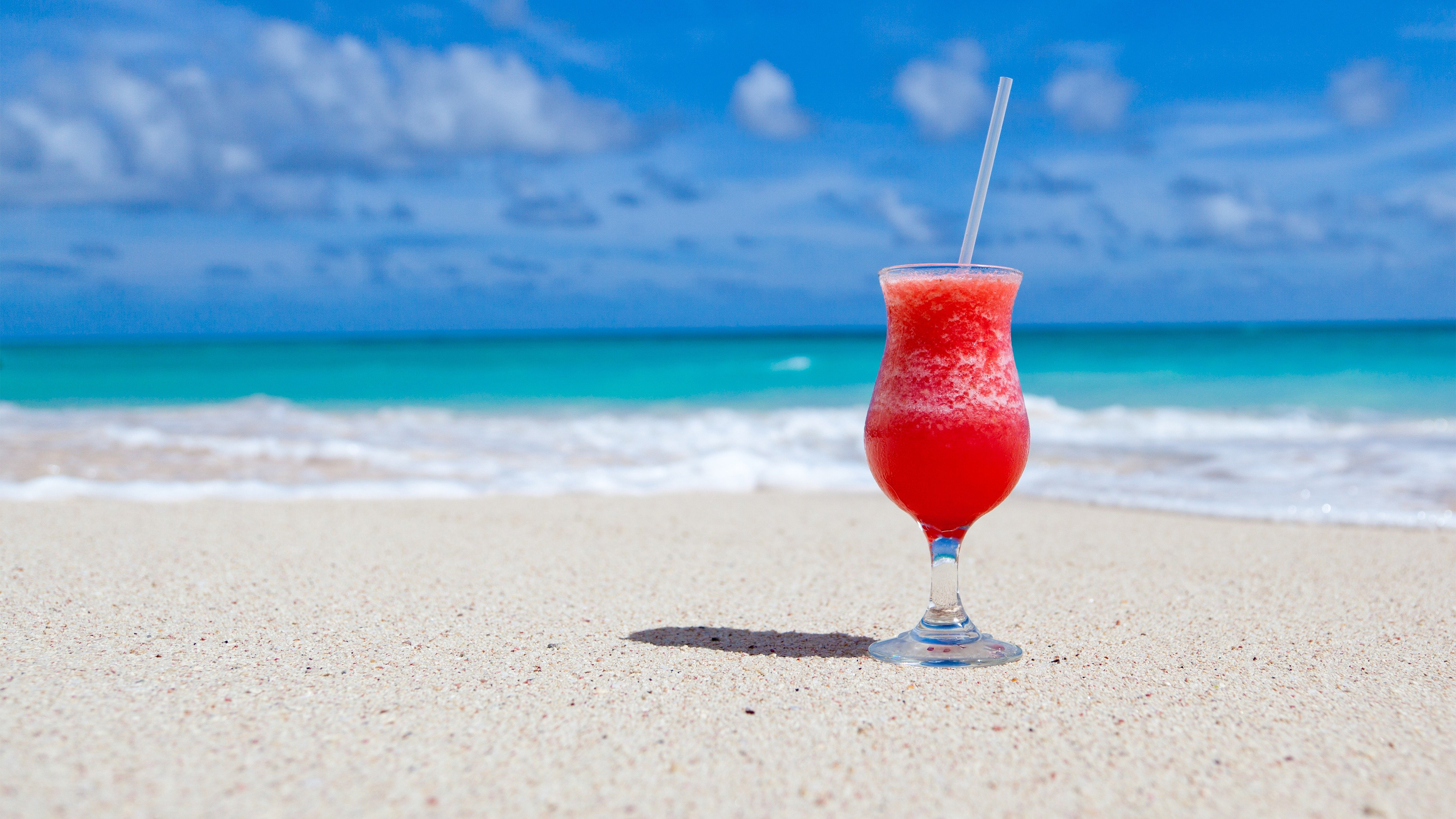}}} & 
\makebox[\linewidth][c]{\raisebox{-0.1\height}{\includegraphics[width=0.9\linewidth]{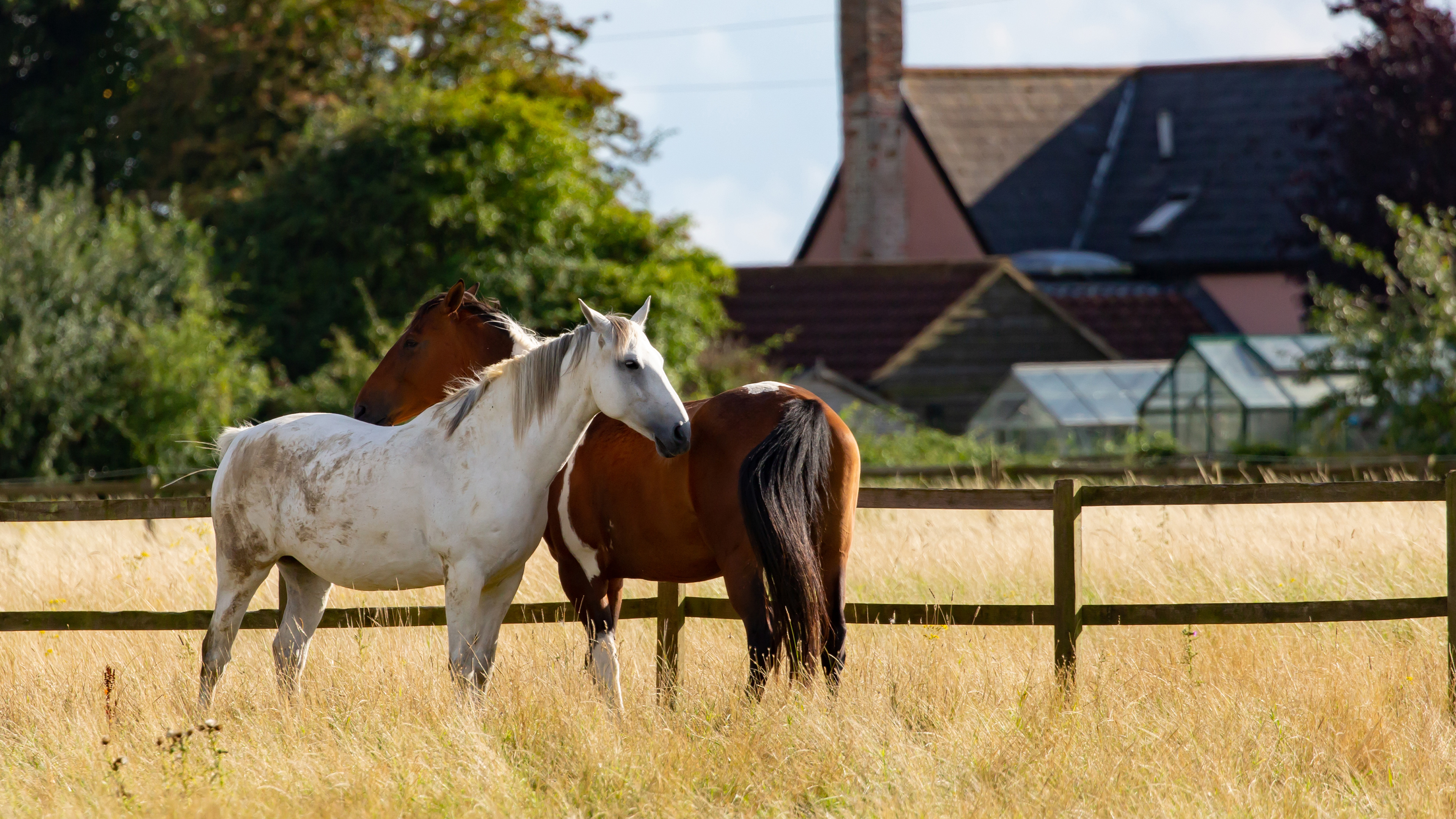}}} & 
\makebox[\linewidth][c]{\raisebox{-0.1\height}{\includegraphics[width=0.5\linewidth]{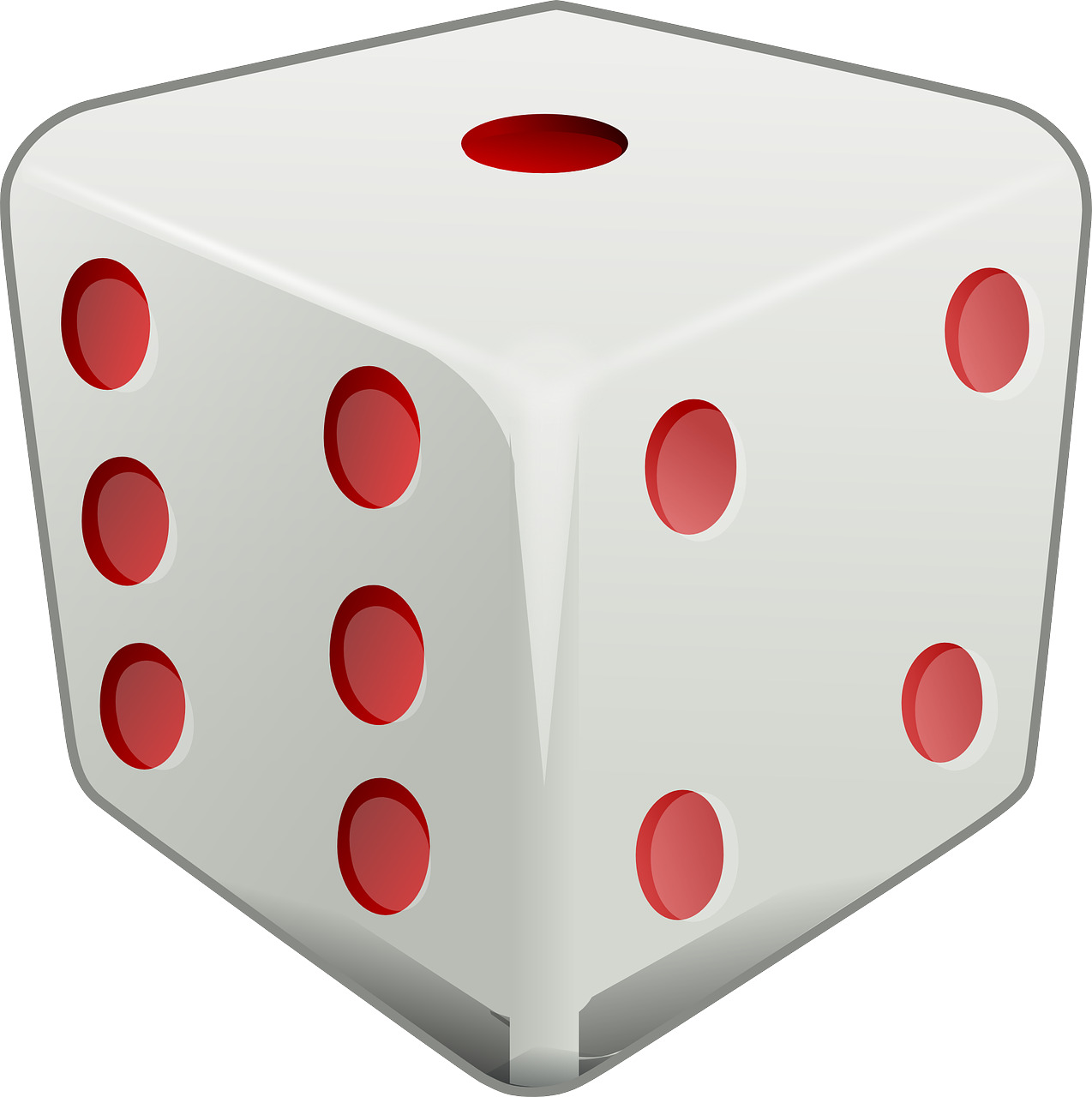}}} & 
\makebox[\linewidth][c]{\raisebox{-0.1\height}{\includegraphics[width=0.75\linewidth]{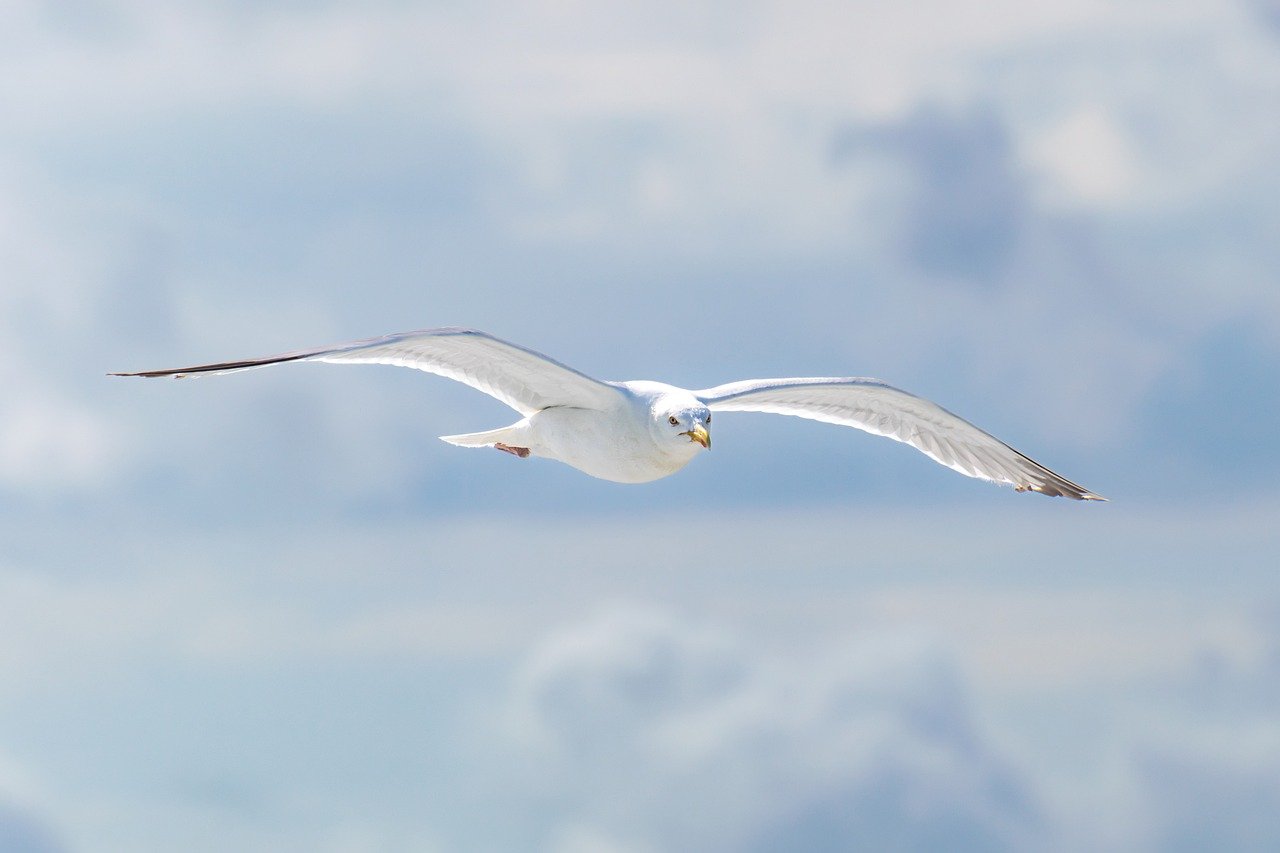}}} \\ \hline
\textbf{\small{Gold Answer}} & 
\multicolumn{1}{>{\raggedright\arraybackslash}m{3.4cm}|}{%
  \begin{small}
  Although it is not visible in the photo, the glass seems to have a back with a shape similar to its front.
  \end{small}
} & 
\multicolumn{1}{>{\raggedright\arraybackslash}m{3.4cm}|}{%
  \begin{small}
  Although it is not visible in the picture, the brown horse's head appears to be connected to the brown horse's torso.
  \end{small}
} & 
\multicolumn{1}{>{\raggedright\arraybackslash}m{3.4cm}|}{%
  \begin{small}
  Although the illustration is drawn on a flat surface, what is depicted appears to be a cube rather than a hexagon.
  \end{small}
} & 
\multicolumn{1}{>{\raggedright\arraybackslash}m{3.4cm}|}{%
  \begin{small}
  Although the photo is flat, it appears that a vast space extends behind the central bird.
  \end{small}
} \\ \hline
\textbf{\small{Miscompletion}} & 
\multicolumn{1}{>{\raggedright\arraybackslash}m{3.4cm}|}{%
  \begin{small}
  Although it is not visible in the photo, the end of the straw appears to be shaped like a spoon.
  \end{small}
} & 
\multicolumn{1}{>{\raggedright\arraybackslash}m{3.4cm}|}{%
  \begin{small}
  Although the photo is flat, the white horse appears to be behind the brown horse.
  \end{small}
} & 
\multicolumn{1}{>{\raggedright\arraybackslash}m{3.4cm}|}{%
  \begin{small}
  Although it is not depicted in the illustration, the opposite side of the 1 face appears to have a 6 face.
  \end{small}
} & 
\multicolumn{1}{>{\raggedright\arraybackslash}m{3.4cm}|}{%
  \begin{small}
  Although they are not visible in the photo, the black spots on the back of the bird in the center appear to be present.
  \end{small}
} \\ \hline
\textbf{\small{No-completion}} & 
\multicolumn{1}{>{\raggedright\arraybackslash}m{3.4cm}|}{%
  \begin{small}
  Although it is not shown in the picture, the end of the straw doesn't appear to continue all the way into the glass.
  \end{small}
} & 
\multicolumn{1}{>{\raggedright\arraybackslash}m{3.4cm}|}{%
  \begin{small}
  Although the photo is flat, the white horse does not appear to be in front of the brown horse.
  \end{small}
} & 
\multicolumn{1}{>{\raggedright\arraybackslash}m{3.4cm}|}{%
  \begin{small}
  Although it is not depicted in the illustration, it doesn't look like there is a white square surface on the opposite side of the eye of 1.
  \end{small}
} & 
\multicolumn{1}{>{\raggedright\arraybackslash}m{3.4cm}|}{%
  \begin{small}
  Although it is not visible in the photo, the bird in the center does not appear to have white feathers on its back.
  \end{small}
} \\ \hline
\textbf{\small{BFO Category}} & 
\begin{small}2-D, Shape\end{small} & 
\begin{small}Part, Color\end{small} & 
\begin{small}Object, Shape\end{small} & 
\begin{small}Site\end{small} \\ \hline
\end{tabular}
\end{table*}

Classified as an L1 category, the \textbf{material entity (Mat)} refers to an independent continuant that includes matter as a constituent. Material entities are three-dimensionally extended and temporally persistent, regardless of the duration of persistence. \textbf{Object (Obj)} and \textbf{object part (Part)} at L2 are subtypes of material entity. A die is an example of an object, whereas the junction between a brown horse's head and its torso serves as an example of an object part. In contrast, the \textbf{immaterial entity (Imm)} is an independent continuant that lacks material constituents. The site and object fiat boundaries are the subtypes of immaterial entities. \textbf{Object fiat boundary (Bou)} is a non-material entity (0D, 1D, or 2D) that does not include a spatial region as a part. Intuitively, it denotes the boundary of a material entity at the interface with its surroundings, and this study considers only \textbf{2D boundaries (2-D)}. Consider question~A in Table~\ref{tab:results_narrow}. The object referenced in the gold-standard answer is the back of the glass. The back of the glass is the surface that exists precisely where the glass meets its surroundings, and the surface itself does not contain any material part. Based on the 2D boundary definition and the correct description, we conclude that the amodally completed object---the back of the glass---is a 2D boundary. Another immaterial entity, \textbf{site (Site)}, is a 3D non-material entity that is either (i) bounded partially or wholly by a material entity or (ii) a 3D non-material part that satisfies condition~(i). For example, in question~D of Table~\ref{tab:question_updated}, the amodally completed entity is a site. Here is our rationale. Predicting the correct answer requires AC of the space behind the bird. Because the space is a 3D non-material entity that is partially bounded by a material entity (i.e., the bird), what is completed in the correct answer is site.\footnote{In fact, the ``space behind the bird'' would move in tandem with the bird. This mobility feature differentiates sites from other immaterial entities, such as spatial regions.}

We now turn from independent continuants to dependent continuants. Dependent continuants are categorized into two primary subtypes: qualities and roles. Because it is controversial to think that \textbf{roles (Rol)} and \textbf{relational qualities (R)} are amodally completed, this study focuses exclusively on qualities. \textbf{Non-relational qualities (NR)} are the typical entities that are considered to be amodally completed. \textbf{Quality (Qua)} is an individual dependent continuant that when it inheres in a continuant is fully realized and instantiated within it. Representative examples of non-relational qualities include \textbf{color (Col)} and \textbf{shape (Sha)}.

Consider again question~A in Table~\ref{tab:question_updated}. If the correct description holds, then it implies that the back of the glass has a specific shape, presumably similar to its front. Thus, by selecting this as the correct answer, we engage in AC of the shape of the glass's back.\footnote{When viewing this image, you might think that the surface continuing to the back of the glass has some specific color. However, the correct answer does not imply that the back of the glass has a specific color. Accordingly, we classify this instance as 2D boundary completion rather than color completion.}

Color and shape are subtypes of qualities but are not formally part of BFO because they are domain-specific categories. For example, there is no room for color in the field of physics.\footnote{Color here should not be conflated with ``color charge'' discussed in physics.} Nevertheless, they are mentioned frequently in existing AC classification studies in the field of computer vision, so we include them in our classification framework. 

\section{Dataset Construction}
\label{sec:exp}

\subsection{Data Collection}
\label{subsec:collection}

This study builds on a dataset from \cite{Nishida2024.07.07.602347} (comprising 69 questions), which was modified to align with our research objectives and extended with additional images and questions. \citet{Nishida2024.07.07.602347} introduced two-choice tasks to assess human understanding of experiences known as horizon consciousness \citep{Gallagher2008-tl}, and because certain aspects of horizon consciousness are related conceptually to AC, selected questions and images were adapted for this study. Image data were mainly sourced from Pixabay,\footnote{\url{https://pixabay.com/}} a royalty-free image repository. 
We added the images and crafted additional answer choices based on three key criteria: (i) part or the whole of the object(s) in the image is occluded in a way that evokes AC, (ii) balanced representation of the categories of completed objects, and (iii) even distribution of image types (drawings vs. photographs).

\subsection{Answer Choices}

For each question, one of the two answer choices was constructed to be easily identifiable as correct by human participants. The other incorrect answer choice was categorized into two types: miscompletion (a description of incorrect completion) and no-completion (a description that fails to involve any completion). Table~\ref{tab:question_updated} shows examples of correct and incorrect answers. Although Table~\ref{tab:question_updated} presents three textual options per image, both model and human evaluations were conducted using two-choice question-answering tasks. Accordingly, two alternative choice pairs were constructed: (gold, miscompletion) and (gold, non-completion). Each question was annotated with up to two BFO categories corresponding to the entity that must be amodally completed to select the correct answer. The number of questions corresponding to each category is indicated in Table~\ref{tab:question_updated}. 

The primary experiment was conducted in Japanese, with an additional model evaluation performed in English. Prompts and answer choices were translated from Japanese into English.

\subsection{Human Evaluation}

Human evaluation was conducted to ensure that the correct answers were easily identifiable by human participants. Only the Japanese version of the prompts (used in model evaluation) was presented during the human evaluation phase.

For each image, we presented participants with either a (gold, miscompletion) or (gold, no-completion) pair of choices and asked them to select the appropriate description based on the given prompt. To avoid bias, the questions were divided into two sets to ensure that each participant viewed only one version of each image--question pair.

In total, 101 responses were collected online during December~6--26, 2024 (63 males, 38 females; age distribution: eight in their teens, 86 in their 20s, six in their 30s, and one in their 40s). Questions with correct response rates below 50\% were excluded as unreliable. As a result, 122 questions were retained as valid: half featured miscompletion choices, and the other half no-completion ones. 

\section{Verification}

\subsection{Models}

We evaluated GPT-4o\footnote{\url{https://openai.com/index/GPT-4o-system-card/}} and Claude 3.5 Sonnet\footnote{\url{https://assets.anthropic.com/m/61e7d27f8c8f5919/original/Claude-3-Model-Card.pdf}} as representative commercial LVLMs.\footnote{We conducted our experiment from December 2024 to January 2025.} Additionally, we assessed three scales of the LLaVA-NeXT series from the Hugging Face Hub---LLaVA-NeXT-34B, LLaVA-NeXT-72B, and LLaVA-NeXT-110B\footnote{\cite{liu2024llavanext,li2024llavanext-72-110}}---to examine whether and how model performance scales with parameter size. Their respective base LLMs are Yi-34B, Qwen1.5-72B, and Qwen1.5-110B. For brevity, we refer to Claude 3.5 Sonnet as Claude, and the LLaVA-NeXT variants as LLaVA-34B, LLaVA-72B, and LLaVA-110B.

\subsection{Evaluation Methods and Metrics}

Zero-shot evaluations were conducted using accuracy as the primary performance metric. We also evaluated the accuracy when blank dummy images were presented in place of the original images to explore whether the models might select plausible answers based solely on text. 

\subsection{Prompts}
\label{sec:prompt} 

Below is an example of the prompts used in the experiment.
\begin{itembox}[c]{Prompts used in experiment}
Instructions: You will see an image and two sentences describing the experience of looking at the image. One of the sentences is correct, and the other is incorrect. Please answer which of the two sentences is appropriate by writing the number of the sentence and only the number.
Explanation 1: [...]
Explanation 2: [...]
\end{itembox}
The complete prompt is obtained by replacing the placeholders ([...]) with textual descriptions, such as those shown in Table~\ref{tab:question_updated}. To avoid biases caused by the order of the descriptions or the labeling of the choices, the answer choices were shuffled randomly. This ensured a roughly equal distribution of correct answers between Explanation~1 and Explanation~2. The same prompts were used for both original and dummy image conditions. This prompt was slightly more detailed than the version used for human evaluation.

\subsection{Results}

First, Table~\ref{tab:total} shows each model's accuracy for the entire dataset. The English result is also displayed. Commercial models such as GPT-4o and Claude approached human-level performance, whereas the LLaVA models showed room for improvement. In English, however, the performance gap between commercial and open models was reduced. 

Next, Table~\ref{tab:results_narrow} shows the accuracy for LVLMs and human participants, grouped by the type of amodally completed object. By employing BFO-based classifications, this categorization revealed specific areas of weakness for each model. For example, GPT-4o showed reduced performance in understanding AC involving object parts (77\%), while Claude performed relatively poorly on color completion tasks (83\%). All LLaVA variants struggled with boundary (56-57\%) and color (53-56\%) completions. Accuracy in color is also lower in GPT-4o (86\%) compared to other categories, suggesting that color-based AC remains a persistent challenge for LVLMs.

\begin{table}[!tb]
\centering
\caption{Accuracy in total. Total accuracy is calculated by the number of questions models could answer correctly divided by the total number of questions. Numbers in parentheses indicate the accuracy in dummy questions, where blank images were provided instead of the original images. Accuracy for humans represents an average value. We did not conduct human experiments with English instructions.}
\label{tab:total}
\resizebox{\columnwidth}{!}{%
\begin{tabular}{@{}ccccccc@{}}
\toprule
Model &
  \begin{tabular}[c]{@{}c@{}}GPT\\ -4o\end{tabular} &
  Claude &
  \begin{tabular}[c]{@{}c@{}}LLaVA\\ -34B\end{tabular} &
  \begin{tabular}[c]{@{}c@{}}LLaVA\\ -72B\end{tabular} &
  \begin{tabular}[c]{@{}c@{}}LLaVA\\ -110B\end{tabular} &
  Human \\ \midrule
\begin{tabular}[c]{@{}c@{}}Accuracy\\ (Japanese)\end{tabular} &
  \begin{tabular}[c]{@{}c@{}}89\\ (77)\end{tabular} &
  \begin{tabular}[c]{@{}c@{}}93\\ (91)\end{tabular} &
  \begin{tabular}[c]{@{}c@{}}56\\ (50)\end{tabular} &
  \begin{tabular}[c]{@{}c@{}}67\\ (54)\end{tabular} &
  \begin{tabular}[c]{@{}c@{}}60\\ (56)\end{tabular} &
  94 \\ \midrule
\begin{tabular}[c]{@{}c@{}}Accuracy\\ (English)\end{tabular} &
  \begin{tabular}[c]{@{}c@{}}88\\ (80)\end{tabular} &
  \begin{tabular}[c]{@{}c@{}}90\\ (88)\end{tabular} &
  \begin{tabular}[c]{@{}c@{}}70\\ (52)\end{tabular} &
  \begin{tabular}[c]{@{}c@{}}79\\ (62)\end{tabular} &
  \begin{tabular}[c]{@{}c@{}}90\\ (78)\end{tabular} &
  - \\ \bottomrule
\end{tabular}%
}
\end{table}

Additional findings from Table~\ref{tab:results_narrow} are summarized below. First, the models achieved relatively high accuracy rates for dummy questions. Interestingly, in some categories and models, the latter achieved higher accuracy with dummy images than with the original images. Claude outperformed its original-image results in two categories when evaluated with dummy images. Second, LLaVA-110B performed worse than the smaller LLaVA-72B. Across all categories, there were instances where LLaVA-72B answered correctly but LLaVA-110B did not. Finally, we analyzed error patterns based on whether the incorrect answer involved miscompletion or no-completion (Table~\ref{tab:error}). A particularly notable finding is that for all the models, the proportion of no-completion errors increased when the models were shown original images compared to dummy images. 

\begin{table}[!t]
\caption{Accuracy by BFO categories in Japanese. Numbers in parentheses indicate the accuracy in dummy questions, where blank images were provided instead of the original images. Cells are shaded gray when the accuracy for dummy questions is higher than for the original questions. For each model, the lowest score is made bold when the original image is presented. The value for the human performance represents the average accuracy. }
\centering
\resizebox{\columnwidth}{!}{%
\begin{tabular}{clcccccccccc}
\hline
L1  &  & \multicolumn{3}{c}{Mat} &  & \multicolumn{3}{c}{Imm} &  & \multicolumn{2}{c}{Qua}      \\ \cline{1-1} \cline{3-5} \cline{7-9} \cline{11-12} 
L2  &  & Obj     &     & Part    &  & Bou     &     & Site    &  & \multicolumn{2}{c}{{[}NR{]}} \\ \cline{1-1} \cline{3-3} \cline{5-5} \cline{7-7} \cline{9-9} \cline{11-12} 
L3  &  &         &     &         &  & 2-D     &     &         &  & {[}Col{]}     & {[}Sha{]}    \\ \cline{1-1} \cline{3-3} \cline{5-5} \cline{7-7} \cline{9-9} \cline{11-12} 
\# questions &  & 16      &     & 38      &  & 54      &     & 18      &  & 36            & 12           \\ \hline \hline 
GPT-4o &
   &
  \begin{tabular}[c]{@{}c@{}}88\\ (75)\end{tabular} &
   &
  \begin{tabular}[c]{@{}c@{}}\textbf{77}\\ (74)\end{tabular} &
   &
  \begin{tabular}[c]{@{}c@{}}94\\ (72)\end{tabular} &
   &
  \begin{tabular}[c]{@{}c@{}}100\\ (100)\end{tabular} &
   &
  \begin{tabular}[c]{@{}c@{}}86\\ (64)\end{tabular} &
  \begin{tabular}[c]{@{}c@{}}92\\ (83)\end{tabular} \\ \hline
Claude &
   &
  \begin{tabular}[c]{@{}c@{}}100\\ (81)\end{tabular} &
   &
  \begin{tabular}[c]{@{}c@{}}90\\ (90)\end{tabular} &
   &
  \begin{tabular}[c]{@{}c@{}}93\\ (93)\end{tabular} &
   &
  \cellcolor[HTML]{EFEFEF}\begin{tabular}[c]{@{}c@{}}94\\ (100)\end{tabular} &
   &
  \cellcolor[HTML]{EFEFEF}\begin{tabular}[c]{@{}c@{}}\textbf{83}\\ (89)\end{tabular} &
  \begin{tabular}[c]{@{}c@{}}100\\ (83)\end{tabular} \\ \hline
\begin{tabular}[c]{@{}c@{}}LLaVA\\ -34B\end{tabular} &
   &
  \begin{tabular}[c]{@{}c@{}}69\\ (62)\end{tabular} &
   &
  \begin{tabular}[c]{@{}c@{}}\textbf{46}\\ (44)\end{tabular} &
   &
  \begin{tabular}[c]{@{}c@{}}56\\ (50)\end{tabular} &
   &
  \begin{tabular}[c]{@{}c@{}}72\\ (61)\end{tabular} &
   &
  \begin{tabular}[c]{@{}c@{}}56\\ (50)\end{tabular} &
  \begin{tabular}[c]{@{}c@{}}75\\ (67)\end{tabular} \\ \hline
\begin{tabular}[c]{@{}c@{}}LLaVA\\ -72B\end{tabular} &
   &
  \begin{tabular}[c]{@{}c@{}}62\\ (50)\end{tabular} &
   &
  \begin{tabular}[c]{@{}c@{}}69\\ (41)\end{tabular} &
   &
  \begin{tabular}[c]{@{}c@{}}57\\ (48)\end{tabular} &
   &
  \begin{tabular}[c]{@{}c@{}}100\\ (100)\end{tabular} &
   &
  \begin{tabular}[c]{@{}c@{}}\textbf{56}\\ (31)\end{tabular} &
  \cellcolor[HTML]{EFEFEF}\begin{tabular}[c]{@{}c@{}}67\\ (75)\end{tabular} \\ \hline
\begin{tabular}[c]{@{}c@{}}LLaVA\\ -110B\end{tabular} &
   &
  \begin{tabular}[c]{@{}c@{}}62 \\ (63)\end{tabular} &
   &
  \begin{tabular}[c]{@{}c@{}}\textbf{51}\\ (49)\end{tabular} &
   &
  \begin{tabular}[c]{@{}c@{}}57\\ (52)\end{tabular} &
   &
  \begin{tabular}[c]{@{}c@{}}89\\ (83)\end{tabular} &
   &
  \begin{tabular}[c]{@{}c@{}}53\\ (50)\end{tabular} &
  \begin{tabular}[c]{@{}c@{}}67\\ (67)\end{tabular} \\ \hline \hline
Human  &  & 95      &     & 94      &  & 94      &     & 92      &  & 92            & 95          \\
\end{tabular}%
}

\label{tab:results_narrow}
\end{table}

\section{Discussion}

The results indicate that each model exhibits specific weaknesses in interpreting texts related to AC, with color completion emerging as a consistent challenge across all models. This finding is critical for deploying LVLM-based systems in real-world scenarios involving diverse types of AC, and it highlights the value of ontological frameworks such as BFO in constructing meaningful and diagnostic  benchmarks.

Furthermore, the findings presented in the previous section raise the following questions. 
\begin{description}[%
  font=\normalfont,       %
  labelwidth=3em,         %
  labelsep=0.1em,         %
  leftmargin=!,           %
]
  \item[Q1)] How do models achieve accuracy above chance on dummy questions?
  \item[Q2)] Why do some models perform better on dummy images than on original images for certain categories?
  \item[Q3)] Why does LLaVA-110B underperform compared to the smaller LLaVA-72B?
\end{description}

We also conducted the same experiments using English prompts. In contrast to the Japanese results, the English experiments showed a more expected trend: model performance improved consistently with larger base LLM sizes, and models generally performed better when original images were presented compared to dummy images (except the ``site'' category). These observations suggest that the abovementioned anomalies are specific to the Japanese experiment. Below, we propose possible explanations for each question.

\paragraph{Hypothesis~1}{Texts affirming the existence of amodal completion appear more frequently in the training dataset than those denying it.}

If the models are biased toward selecting gold or miscompletion answers over no-completion alternatives because of the distribution of training data, then this could explain their tendency to perform above chance (50\%) even when presented with dummy images. 

\paragraph{Hypothesis~2} \textbf{The models have visual limitations in amodal completion.}

How can Q2 and Q3 be addressed? One possible explanation centers on the models' visual capability. While LVLMs may select answers correctly based on textual cues alone, they may choose incorrectly when the original image is presented because of limited visual competence in AC. The textual input may assign higher probability to the correct answer, but the visual input could introduce conflicting cues, thereby lowering accuracy. Although this hypothesis is plausible, it does not rule out other contributing factors, which we explore below.

\begin{table}[!tb]
\centering
\caption{Error analysis. The number of incorrect answers by a pattern of erroneous choices. The percentage of no-completions among all incorrectly answered questions increases when the original image is presented. Numbers in parenthesis indicate the accuracy in dummy questions.}
\label{tab:error}
\resizebox{\columnwidth}{!}{%
\begin{tabular}{@{}cccc@{}}
\toprule
 & Miscompletion & No-completion & \begin{tabular}[c]{@{}c@{}}\% no-completion \\ among errors\end{tabular} \\ \midrule
GPT-4o     & \begin{tabular}[c]{@{}c@{}}3\\ (7)\end{tabular}   & \begin{tabular}[c]{@{}c@{}}10\\ (21)\end{tabular}  & \begin{tabular}[c]{@{}c@{}}77\%\\ (75\%)\end{tabular} \\ \midrule
Claude     & \begin{tabular}[c]{@{}c@{}}4\\ (8)\end{tabular}   & \begin{tabular}[c]{@{}c@{}}5\\ (3)\end{tabular}   & \begin{tabular}[c]{@{}c@{}}56\%\\ (27\%)\end{tabular}  \\ \midrule
LLaVA-34B  & \begin{tabular}[c]{@{}c@{}}25\\ (30)\end{tabular}   & \begin{tabular}[c]{@{}c@{}}29\\ (31)\end{tabular}   & \begin{tabular}[c]{@{}c@{}}54\%\\ (51\%)\end{tabular}   \\ \midrule
LLaVA-72B  & \begin{tabular}[c]{@{}c@{}}17\\ (27)\end{tabular}  & \begin{tabular}[c]{@{}c@{}}23\\ (30)\end{tabular}  & \begin{tabular}[c]{@{}c@{}}\%57\\ (\%53)\end{tabular}  \\ \midrule
LLaVA-110B & \begin{tabular}[c]{@{}c@{}}24\\ (30)\end{tabular}  & \begin{tabular}[c]{@{}c@{}}25\\ (24)\end{tabular}  & \begin{tabular}[c]{@{}c@{}}51\%\\ (44\%)\end{tabular} \\ \bottomrule
\end{tabular}%
}
\end{table}

\paragraph{Hypothesis~3} \textbf{The models do not recognize the evidential function of Japanese perceptual verbs.}

Hypothesis~3 answers Q2 and Q3 from the perspective of the models' language understanding capabilities. Because the task involves multimodal inputs, linguistic comprehension plays a crucial role in performance. In particular, the models may struggle to interpret the perceptual verb \textit{mieru}, which appears in every correct description of AC in our dataset. While the literal meaning of \textit{mieru} is ``can be seen,'' according to \citet{shiba2023}, the verb \textit{mieru} undergoes what is called ``grammaticalization,'' a phenomenon in which words that have substantive meaning and can function as independent elements, known as ``content words,'' change into words that have the character of ``function words,'' which are elements that solely serve grammatical functions \citep{Miyake_2005_grammaticalization}. \citet{shiba2023} points out, when combined with the evidential morpheme \textit{yooda} to form \textit{yoo-ni mieru}, it functions as an evidential marker, indicating that the speaker's statement is based on visual evidence. In this evidential usage, native Japanese speakers select the correct answer easily because the marker permits indirect evidence, making it compatible with the fact that the object in question is not directly visible.

Hypothesis~3 may answer Q2---why some models perform better on dummy images than on original images for certain categories---by suggesting that some LVLMs fail to recognize the evidential use of \textit{mieru}. As a result, models may avoid descriptions that (to them) imply an invisible object ``can be seen,'' even if they are capable of amodally completing the occluded parts of the image. This tendency to select the no-completion option is manifested in the fact that all the models choose no-completion options more when they are presented with the original image than with the dummies (Table \ref{tab:error}). 

It may also answer Q3, namely, why LLaVA-72B outperformed the larger LLaVA-110B. According to \citep{li2024llavanext-72-110}, LLaVA-110B performs better than 72B on multimodal benchmarks. This suggests that LLaVA-110B may be more sensitive to visual input. However, if it still struggles to interpret the evidential meaning of \textit{mieru}, then the model may be more hesitant to select the correct answer, particularly when textual cues hinge on this subtle linguistic distinction.

Evidentiality manifests differently across languages. For example, Japanese has both grammaticalized and non-grammaticalized evidential forms, whereas English lacks grammaticalized evidentials, relying instead on verbs and adverbs such as ``seem'' or ``appear'' to express similar functions \citep{Yang2014-xt}. Given the findings of the present study, the diversity of evidential expressions across languages may pose a considerable challenge for LVLMs. Because evidentiality is a fundamental linguistic mechanism for expressing subjective perceptual experiences including AC, its cross-linguistic variability represents an important direction for future research in improving the multilingual reasoning and generalization of LVLMs.

\section{Conclusion}
\label{sec:conc}

In this study, we evaluated the multimodal capabilities of LVLMs on tasks involving both text and images related to AC. Rather than focusing solely on image processing, our objective was to assess systematically how models comprehended texts describing AC by using a classification framework derived from BFO.

The results showed that while LVLMs demonstrate a generally strong understanding of AC-related texts (comparable to human performance), their accuracy varies considerably depending on the type of object being completed, in contrast to the relatively stable performance observed in human participants. Moreover, experiments using dummy images revealed an intriguing language-specific effect: in Japanese, models such as LLaVA-NeXT and Claude sometimes performed worse with original images than with dummy images. This pattern suggests a limited understanding of Japanese perceptual verbs, which are essential for the correct interpretation of AC-related descriptions.

In future work, we will investigate further the hypothesis that these models fail to grasp the evidential use of perceptual verbs in Japanese. Advancing this line of research may enhance the ability of LVLMs to interpret human-like descriptions of amodal perception and support applications in the automated analysis of subjective narratives related to conscious experience.

\clearpage

\section{Acknowledgments}
This work was supported by JSPS KAKENHI grant number JP24H00809, 24K22328, 24KK0189, and 23K00001.

\bibliographystyle{apacite}

\setlength{\bibleftmargin}{.125in}
\setlength{\bibindent}{-\bibleftmargin}

\bibliography{CogSci}

\begin{thebibliography}{}

\bibitem [\protect \citeauthoryear {%
Ao%
, Ke%
\BCBL {}\ \BBA {} Ehinger%
}{%
Ao%
\ \protect \BOthers {.}}{%
{\protect \APACyear {2023}}%
}]{%
AO-2023}
\APACinsertmetastar {%
AO-2023}%
\begin{APACrefauthors}%
Ao, J.%
, Ke, Q.%
\BCBL {}\ \BBA {} Ehinger, K\BPBI A.%
\end{APACrefauthors}%
\unskip\
\newblock
\APACrefYearMonthDay{2023}{}{}.
\newblock
{\BBOQ}\APACrefatitle {Image amodal completion: A survey} {Image amodal completion: A survey}.{\BBCQ}
\newblock
\APACjournalVolNumPages{Computer Vision and Image Understanding}{229}{}{103661}.
\newblock
\begin{APACrefDOI} \doi{10.1016/j.cviu.2023.103661} \end{APACrefDOI}
\PrintBackRefs{\CurrentBib}

\bibitem [\protect \citeauthoryear {%
Arp%
, Smith%
\BCBL {}\ \BBA {} Spear%
}{%
Arp%
\ \protect \BOthers {.}}{%
{\protect \APACyear {2015}}%
}]{%
Arp2015-rt}
\APACinsertmetastar {%
Arp2015-rt}%
\begin{APACrefauthors}%
Arp, R.%
, Smith, B.%
\BCBL {}\ \BBA {} Spear, A\BPBI D.%
\end{APACrefauthors}%
\unskip\
\newblock
\APACrefYear{2015}.
\newblock
\APACrefbtitle {Building Ontologies with Basic Formal Ontology} {Building ontologies with basic formal ontology}.
\newblock
\APACaddressPublisher{London, England}{MIT Press}.
\PrintBackRefs{\CurrentBib}

\bibitem [\protect \citeauthoryear {%
Gallagher%
\ \BBA {} Zahavi%
}{%
Gallagher%
\ \BBA {} Zahavi%
}{%
{\protect \APACyear {2008}}%
}]{%
Gallagher2008-tl}
\APACinsertmetastar {%
Gallagher2008-tl}%
\begin{APACrefauthors}%
Gallagher, S.%
\BCBT {}\ \BBA {} Zahavi, D.%
\end{APACrefauthors}%
\unskip\
\newblock
\APACrefYear{2008}.
\newblock
\APACrefbtitle {Phenomenological Mind: An Introduction to Philosophy of Mind and Cognitive Science} {Phenomenological mind: An introduction to philosophy of mind and cognitive science}.
\newblock
\APACaddressPublisher{New York, NY}{Routledge}.
\PrintBackRefs{\CurrentBib}

\bibitem [\protect \citeauthoryear {%
Gerbino%
}{%
Gerbino%
}{%
{\protect \APACyear {2020}}%
}]{%
Gerbino2020-yk}
\APACinsertmetastar {%
Gerbino2020-yk}%
\begin{APACrefauthors}%
Gerbino, W.%
\end{APACrefauthors}%
\unskip\
\newblock
\APACrefYearMonthDay{2020}{}{}.
\newblock
{\BBOQ}\APACrefatitle {Amodal completion revisited} {Amodal completion revisited}.{\BBCQ}
\newblock
\APACjournalVolNumPages{i-Perception}{11}{4}{2041669520937323}.
\newblock
\begin{APACrefDOI} \doi{10.1177/2041669520937323} \end{APACrefDOI}
\PrintBackRefs{\CurrentBib}

\bibitem [\protect \citeauthoryear {%
Li%
\ \protect \BOthers {.}}{%
Li%
\ \protect \BOthers {.}}{%
{\protect \APACyear {2024}}%
}]{%
li2024llavanext-72-110}
\APACinsertmetastar {%
li2024llavanext-72-110}%
\begin{APACrefauthors}%
Li, B.%
, Zhang, K.%
, Zhang, H.%
, Guo, D.%
, Zhang, R.%
, Li, F.%
\BDBL {}Li, C.%
\end{APACrefauthors}%
\unskip\
\newblock
\APACrefYearMonthDay{2024}{}{}.
\newblock
\APACrefbtitle {{LLaVA-NeXT: Stronger LLMs Supercharge Multimodal Capabilities in the Wild}.} {{LLaVA-NeXT: Stronger LLMs Supercharge Multimodal Capabilities in the Wild}.}
\newblock
\begin{APACrefURL} [{2025-4-25}]\url{https://llava-vl.github.io/blog/2024-05-10-llava-next-stronger-llms/} \end{APACrefURL}
\PrintBackRefs{\CurrentBib}

\bibitem [\protect \citeauthoryear {%
Liu%
\ \protect \BOthers {.}}{%
Liu%
\ \protect \BOthers {.}}{%
{\protect \APACyear {2024}}%
}]{%
liu2024llavanext}
\APACinsertmetastar {%
liu2024llavanext}%
\begin{APACrefauthors}%
Liu, H.%
, Li, C.%
, Li, Y.%
, Li, B.%
, Zhang, Y.%
, Shen, S.%
\BCBL {}\ \BBA {} Lee, Y\BPBI J.%
\end{APACrefauthors}%
\unskip\
\newblock
\APACrefYearMonthDay{2024}{}{}.
\newblock
\APACrefbtitle {{LLaVA-NeXT: Improved reasoning, OCR, and world knowledge}.} {{LLaVA-NeXT: Improved reasoning, OCR, and world knowledge}.}
\newblock
\begin{APACrefURL} \url{https://llava-vl.github.io/blog/2024-01-30-llava-next/} \end{APACrefURL}
\PrintBackRefs{\CurrentBib}

\bibitem [\protect \citeauthoryear {%
Michotte%
}{%
Michotte%
}{%
{\protect \APACyear {1991}}%
}]{%
Michotte1991-kp}
\APACinsertmetastar {%
Michotte1991-kp}%
\begin{APACrefauthors}%
Michotte, A.%
\end{APACrefauthors}%
\unskip\
\newblock
\APACrefYearMonthDay{1991}{}{}.
\newblock
{\BBOQ}\APACrefatitle {Amodal completion of perceptual structures} {Amodal completion of perceptual structures}.{\BBCQ}
\newblock
\BIn{} G.~Thinés, A.~Costall\BCBL {}\ \BBA {} G.~Butterworth\ (\BEDS), \APACrefbtitle {Michotte's experimental phenomenology of perception} {Michotte's experimental phenomenology of perception}\ (\BPGS\ 140--167).
\newblock
\APACaddressPublisher{Hillsdale, N.J.}{Lawrence Erlbaum, Associates}.
\PrintBackRefs{\CurrentBib}

\bibitem [\protect \citeauthoryear {%
Miyake%
}{%
Miyake%
}{%
{\protect \APACyear {2005}}%
}]{%
Miyake_2005_grammaticalization}
\APACinsertmetastar {%
Miyake_2005_grammaticalization}%
\begin{APACrefauthors}%
Miyake, T.%
\end{APACrefauthors}%
\unskip\
\newblock
\APACrefYearMonthDay{2005}{}{}.
\newblock
{\BBOQ}\APACrefatitle {Gendai Nihongo ni Okeru Bunpōka: Naiyōgo to Kinōgo no Renzokusei o Megutte [Grammaticalization in Modern Japanese: On the Continuity of Content Words and Functional Words]} {Gendai nihongo ni okeru bunpōka: Naiyōgo to kinōgo no renzokusei o megutte [grammaticalization in modern japanese: On the continuity of content words and functional words]}.{\BBCQ}
\newblock
\APACjournalVolNumPages{Nihongo no Kenkyū}{1}{3}{61--76}.
\newblock
\begin{APACrefDOI} \doi{10.20666/nihongonokenkyu.1.3\_61} \end{APACrefDOI}
\PrintBackRefs{\CurrentBib}

\bibitem [\protect \citeauthoryear {%
Nanay%
}{%
Nanay%
}{%
{\protect \APACyear {2018}}%
}]{%
Nanay2018-vq}
\APACinsertmetastar {%
Nanay2018-vq}%
\begin{APACrefauthors}%
Nanay, B.%
\end{APACrefauthors}%
\unskip\
\newblock
\APACrefYearMonthDay{2018}{}{}.
\newblock
{\BBOQ}\APACrefatitle {The importance of amodal completion in everyday perception} {The importance of amodal completion in everyday perception}.{\BBCQ}
\newblock
\APACjournalVolNumPages{i-Perception}{9}{4}{1--16}.
\newblock
\begin{APACrefDOI} \doi{10.1177/2041669518788887} \end{APACrefDOI}
\PrintBackRefs{\CurrentBib}

\bibitem [\protect \citeauthoryear {%
Nishida%
, Hamada%
, Niikawa%
\BCBL {}\ \BBA {} Miyahara%
}{%
Nishida%
\ \protect \BOthers {.}}{%
{\protect \APACyear {2024}}%
}]{%
Nishida2024.07.07.602347}
\APACinsertmetastar {%
Nishida2024.07.07.602347}%
\begin{APACrefauthors}%
Nishida, S.%
, Hamada, H\BPBI T.%
, Niikawa, T.%
\BCBL {}\ \BBA {} Miyahara, K.%
\end{APACrefauthors}%
\unskip\
\newblock
\APACrefYearMonthDay{2024}{}{}.
\newblock
{\BBOQ}\APACrefatitle {Neural correlates of phenomenological attitude toward perceptual experience} {Neural correlates of phenomenological attitude toward perceptual experience}.{\BBCQ}
\newblock
\APACjournalVolNumPages{bioRxiv}{}{}{}.
\newblock
\begin{APACrefURL} \url{https://www.biorxiv.org/content/early/2024/07/10/2024.07.07.602347} \end{APACrefURL}
\newblock
\begin{APACrefDOI} \doi{10.1101/2024.07.07.602347} \end{APACrefDOI}
\PrintBackRefs{\CurrentBib}

\bibitem [\protect \citeauthoryear {%
Pessoa%
, Thompson%
\BCBL {}\ \BBA {} Noë%
}{%
Pessoa%
\ \protect \BOthers {.}}{%
{\protect \APACyear {2001}}%
}]{%
Pessoa2001-ys}
\APACinsertmetastar {%
Pessoa2001-ys}%
\begin{APACrefauthors}%
Pessoa, L.%
, Thompson, E.%
\BCBL {}\ \BBA {} Noë, A.%
\end{APACrefauthors}%
\unskip\
\newblock
\APACrefYearMonthDay{2001}{}{}.
\newblock
{\BBOQ}\APACrefatitle {Filling-in: One or many?} {Filling-in: One or many?}{\BBCQ}
\newblock
\APACjournalVolNumPages{Behavioral and Brain Sciences}{24}{6}{1137--1139}.
\newblock
\begin{APACrefDOI} \doi{10.1017/S0140525X01230143} \end{APACrefDOI}
\PrintBackRefs{\CurrentBib}

\bibitem [\protect \citeauthoryear {%
Shiba%
}{%
Shiba%
}{%
{\protect \APACyear {2023}}%
}]{%
shiba2023}
\APACinsertmetastar {%
shiba2023}%
\begin{APACrefauthors}%
Shiba, A.%
\end{APACrefauthors}%
\unskip\
\newblock
\APACrefYearMonthDay{2023}{}{}.
\newblock
{\BBOQ}\APACrefatitle {Chikaku Dōshi ``Mieru'' no Suitei Kōbun e no Hirogari: Kyōjitai ni Okeru Bunpōka [Extension of the Japanese Perception Verb Mieru to the Evidential Construction]} {Chikaku dōshi ``mieru'' no suitei kōbun e no hirogari: Kyōjitai ni okeru bunpōka [extension of the japanese perception verb mieru to the evidential construction]}.{\BBCQ}
\newblock
\APACjournalVolNumPages{The Journal of Humanities, Nagoya University}{6}{}{57--78}.
\newblock
\begin{APACrefDOI} \doi{10.18999/jouhunu.6.57} \end{APACrefDOI}
\PrintBackRefs{\CurrentBib}

\bibitem [\protect \citeauthoryear {%
Tse%
}{%
Tse%
}{%
{\protect \APACyear {1999}}%
}]{%
Tse1999-xi}
\APACinsertmetastar {%
Tse1999-xi}%
\begin{APACrefauthors}%
Tse, P\BPBI U.%
\end{APACrefauthors}%
\unskip\
\newblock
\APACrefYearMonthDay{1999}{}{}.
\newblock
{\BBOQ}\APACrefatitle {Volume completion} {Volume completion}.{\BBCQ}
\newblock
\APACjournalVolNumPages{Cognitive Psychology}{39}{1}{37--68}.
\newblock
\begin{APACrefDOI} \doi{10.1006/cogp.1999.0715} \end{APACrefDOI}
\PrintBackRefs{\CurrentBib}

\bibitem [\protect \citeauthoryear {%
Van~Lier%
\ \BBA {} Gerbino%
}{%
Van~Lier%
\ \BBA {} Gerbino%
}{%
{\protect \APACyear {2015}}%
}]{%
van-Lier2015-gq}
\APACinsertmetastar {%
van-Lier2015-gq}%
\begin{APACrefauthors}%
Van~Lier, R.%
\BCBT {}\ \BBA {} Gerbino, W.%
\end{APACrefauthors}%
\unskip\
\newblock
\APACrefYearMonthDay{2015}{}{}.
\newblock
{\BBOQ}\APACrefatitle {Perceptual completions} {Perceptual completions}.{\BBCQ}
\newblock
\BIn{} \APACrefbtitle {The Oxford Handbook of Perceptual Organization.} {The oxford handbook of perceptual organization.}
\newblock
\APACaddressPublisher{}{Oxford University Press}.
\newblock
\begin{APACrefDOI} \doi{10.1093/oxfordhb/9780199686858.013.040} \end{APACrefDOI}
\PrintBackRefs{\CurrentBib}

\bibitem [\protect \citeauthoryear {%
Yang%
}{%
Yang%
}{%
{\protect \APACyear {2014}}%
}]{%
Yang2014-xt}
\APACinsertmetastar {%
Yang2014-xt}%
\begin{APACrefauthors}%
Yang, L.%
\end{APACrefauthors}%
\unskip\
\newblock
\APACrefYearMonthDay{2014}{}{}.
\newblock
{\BBOQ}\APACrefatitle {Evidentiality in English research articles of applied linguistics: From the perspective of metadiscourse} {Evidentiality in english research articles of applied linguistics: From the perspective of metadiscourse}.{\BBCQ}
\newblock
\APACjournalVolNumPages{Journal of Language Teaching and Research}{5}{3}{581--591}.
\newblock
\begin{APACrefDOI} \doi{10.4304/jltr.5.3.581-591} \end{APACrefDOI}
\PrintBackRefs{\CurrentBib}

\end{thebibliography}

\end{document}